%% file: main.tex
\documentclass[AMS,STIX1COL]{WileyNJD-v2}

\usepackage{amsmath,amsfonts}
\usepackage{bm,subcaption,enumitem,pifont,mathtools,stix,graphicx}
\usepackage{comment}
\usepackage{natbib}

\makeatletter
\let\oldabs\abs
\def\abs{\@ifstar{\oldabs}{\oldabs*}}

\makeatletter
\newif\if@restonecol
\makeatother

\DeclareUnicodeCharacter{2212}{-}

\articletype{Journal}%

\begin{document}

\title{Large Deviations for Accelerating Neural Networks Training}

\author[1,4]{Sreelekha Guggilam*}
\author[1,2]{Varun Chandola}
\author[3]{Abani Patra}
\authormark{GUGGILAM \textsc{et al.}}
\address[4]{\orgdiv{National Security and Science Directorate}, \orgname{Oak Ridge National Lab}, \orgaddress{\state{MA}, \country{USA}}}
\address[1]{\orgdiv{Computational Data Science \& Eng.}, \orgname{University at Buffalo, State University of New York (SUNY)}, \orgaddress{\state{NY}, \country{USA}}}
\address[2]{\orgdiv{Computer Science \& Eng.}, \orgname{University at Buffalo, State University of New York (SUNY)}, \orgaddress{\state{NY}, \country{USA}}}
\address[3]{\orgdiv{Data Intensive Studies Center }, \orgname{Tufts University}, \orgaddress{\state{MA}, \country{USA}}}
\corres{\email{\{guggilams@ornl.gov, chandola@buffalo.edu, abani.patra@tufts.edu\}}}

\abstract[Summary]{
    Artificial neural networks (ANNs) require tremendous amount of data to train on. However, in classification models, most data features are often similar which can lead to increase in training time without significant improvement in the performance. Thus, we hypothesize that there could be a more efficient way to train an ANN using a better representative sample. For this, we propose the LAD Improved Iterative Training (LIIT), a novel training approach for ANN using large deviations principle to generate and iteratively update training samples in a fast and efficient setting. This is exploratory work with extensive opportunities for future work. The thesis presents this ongoing research work with the following contributions from this study:
    $(1)$ We propose a novel ANN training method, LIIT, based on the large deviations theory where additional dimensionality reduction is not needed to study high dimensional data.
    $(2)$ The LIIT approach uses a Modified Training Sample (MTS) that is generated and iteratively updated using a LAD anomaly score based sampling strategy.
    $(3)$ The MTS sample is designed to be well representative of the training data by including most anomalous of the observations in each class. This ensures distinct patterns and features are learnt with smaller samples.
    $(4)$ We study the classification performance of the LIIT trained ANNs with traditional batch trained counterparts. 
}


\maketitle

\input{LADNN/introduction}
\input{LADNN/lit_review_nn}
\input{LADNN/methodology}
\input{LADNN/results}
\section{Acknowledgements}
The authors would like to acknowledge University at Buffalo Center for Computational Research (http://www.buffalo.edu/ccr.html) for its computing resources that were made available for conducting the research reported in this paper. Financial support of the National Science Foundation Grant numbers NSF/OAC 1339765 and NSF/DMS 1621853 is acknowledged.

\bibliography{main}    
\end{document}

%% file: LADNN/introduction.tex
\section{Introduction}
Artificial Neural Networks (ANNs) are assumption free models that gather information from the provided training data. Due to their design, they are ideal to study complex functional dependencies between input and output layers. In contrast to traditional statistical models that use metrics like mean, covariance matrices, probability and confidence intervals, ANNs rely on patterns  in training data for model development and fitting. Though this can be considered useful to develop better fitting architectures than their statistical counterparts, specially in datasets with complex structure, the effect of incorrect or deficient training data is profound.

In this paper, we propose a statistically enhanced sampling of the training data in combination with a novel training method that ensures faster training of the neural network. The following are the contributions of this research:
\begin{enumerate}
    \item We  propose the LAD Improved Iterative Training (LIIT) strategy that uses a Modified Training Sample (MTS) to train the neural network. 
    \item We present four LAD score based sampling strategies to design the MTS. Obtaining the LAD score based on a large deviations principle   is computationally inexpensive. Therefore, one can analyze large and high dimensional datasets without additional dimensionality reduction procedures  allowing more accurate and cost effective scoring schema. 
    \item The use of MTS which is a smaller training sample reduces the cost of computational time significantly for large datasets.
    \item We perform an empirical study on publicly available classification benchmark datasets to analyze the performance of the proposed method.
\end{enumerate}

The work presented here is limited to simple classification based neural networks. Future work will include extending it to more complex ANNs. 


%% file: LADNN/lit_review_nn.tex
\section{Related Work}\label{lit_review_liit}
In this section, we provide a brief overview of sensitivity to training samples and speed of neural network training. 

Artificial neural networks are powerful for general classification. However, its excellent performance often depends largely on a huge training set. A large body of research exists that study the impact of training data size on neural network learning \cite{soekhoe2016impact, djolonga2021robustness}. In particular, it is evident that smaller training data leads to less efficient models. However, the vast computational expense associated with training on large sets of data makes the need to improve training practices essential, specially for online or real-time models.

Many methods that try to model faster neural networks exist. For instance, \citet{wang2019batch} use batch normalization in deep neural networks to improve the convergence rates. \citet{zhong2017satcnn} work on image classification using their agile convolution neural network SatCNN, for quick and effective learning with small convolutional kernels and deep convolutional layers. However, these works are limited to the domain problems and cannot be easily scaled to other data types. 

Another alternative to improve the training speed can be by modifying the training samples. For instance, studies like \citet{shanker1996effect} look at the effect of standardization of data on the learning of the neural network. \citet{kavzoglu2009increasing} emphasizes on characteristics of training samples and uses representative training to improve the classification. These methods, however, fail to study the impact of smaller data on model performance and efficiency. 

In this part of the thesis, we propose a novel training strategy that can be generalized across domains. The method is used to replicate the true representation of the training features in a smaller sample which can be in turn used for faster training and convergence. Due to the proper representation of even the most extreme observations, this method ensures faster learning with competitive performance. 

%% file: LADNN/methodology.tex
\section{Methodology}\label{methodology_liit}
The most important aspect of classification models is the adequacy of the representative training samples for each class. Although the size of the training data is of considerable importance, acquiring a large number of representative training data may be impractical where a large number of classes are involved. In particular, since most observations within each true class have similar features, multiple samples add low value in terms of novel information/pattern. In this section, we describe the traditional batch training approach in brief followed by the LAD Improved Iterative Training approach. We present 4 sampling strategies used in the LIIT training and their respective algorithms. 

\input{LADNN/definitions}

\subsection{Classification Neural Network}
\begin{figure}
     \centering
         \centering
         \includegraphics[width=0.49\textwidth]{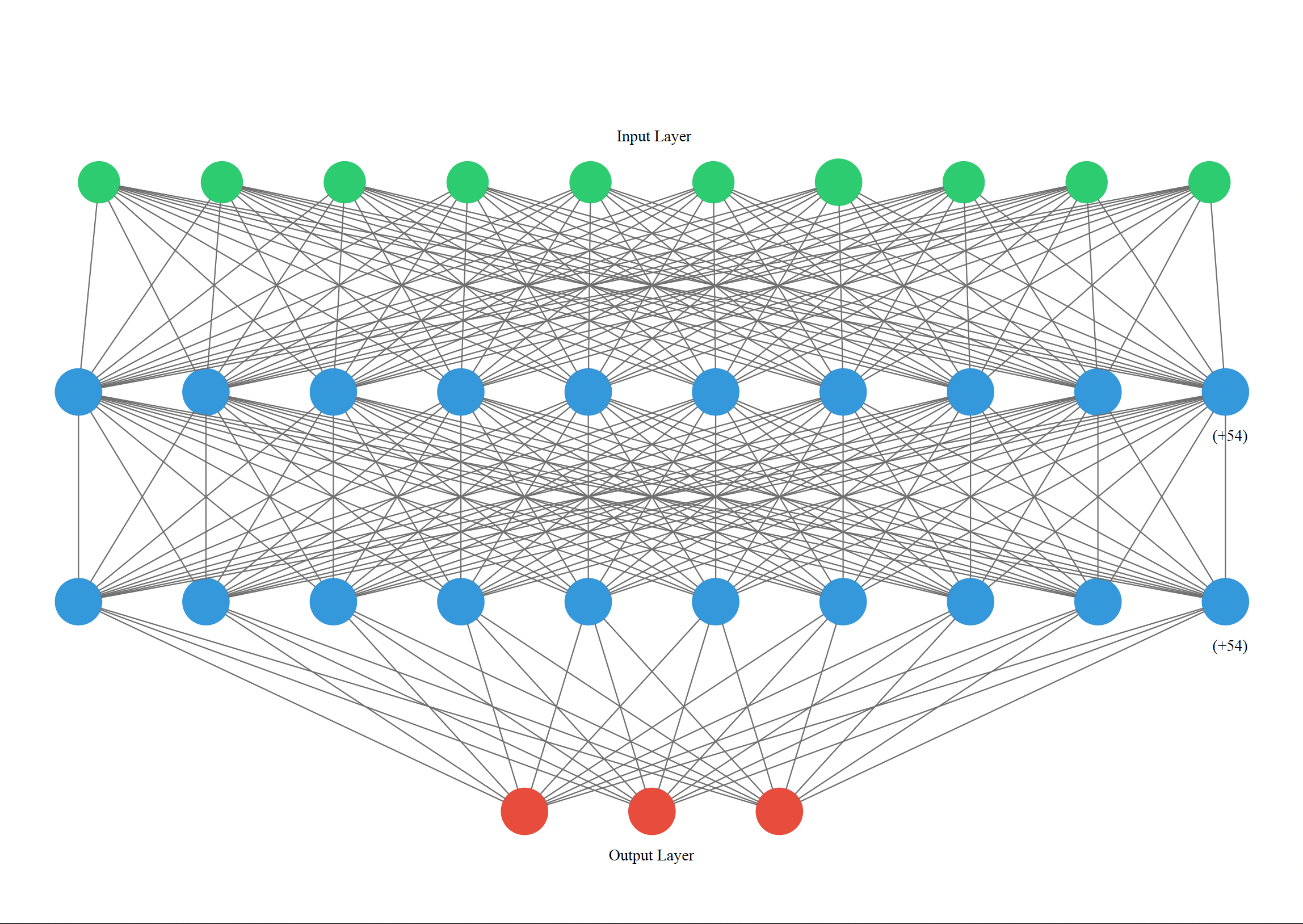}
         \caption{Simple Classification Neural Network: The figure illustrates a dense neural network to classify data into 3 classes. The network takes an input of 10 dimensions and returns scores used to assign each class.}
         \label{fig:nn_model}
\end{figure}

For this analysis, we look at a   basic classification algorithm. Figure \ref{fig:nn_model} shows the architecture of a simple three layer dense neural network. 

The model is trained using full training samples with the convergence criterion set to zero validation loss for 5 epochs with the maximum number of epochs   set to 180. Three different activation functions, RELU, Tanh, Softmax are used for the three consecutive dense layers respectively. 
A simple model was chosen to study the proof of concept of the representative sampling strategy presented in the part of the thesis. Further studies are needed to understand the relation between the model choice and training sampling techniques. 


\begin{figure*}
    \centering
    \includegraphics[width=0.95\textwidth]{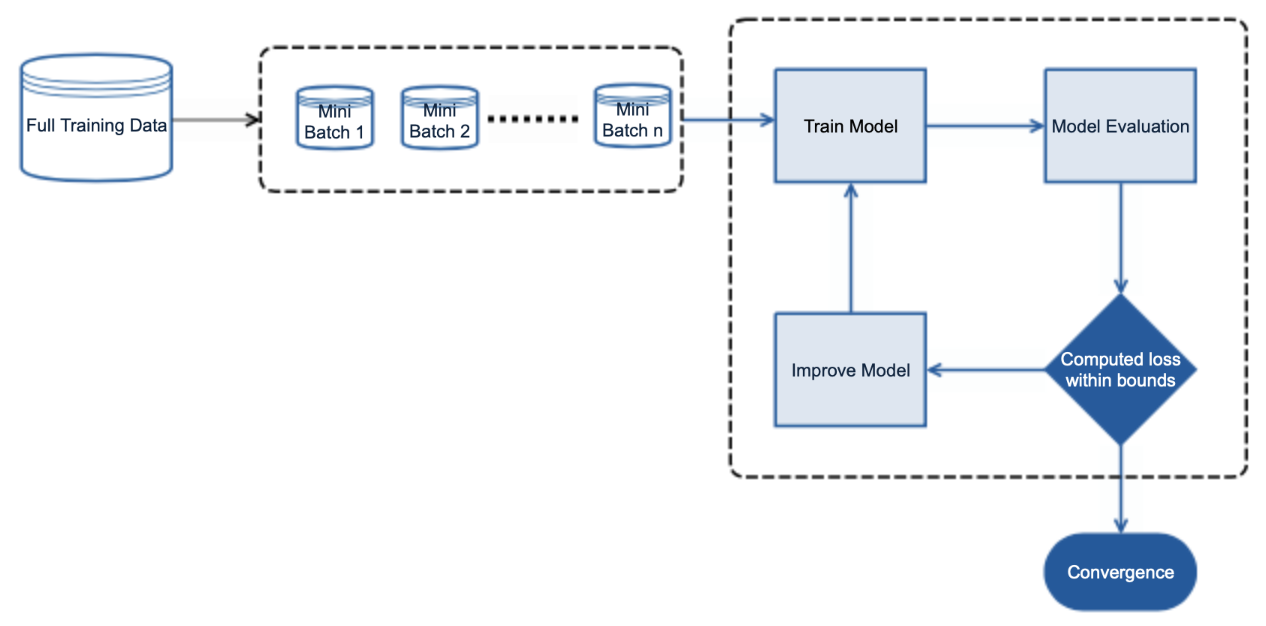}
    \caption{Mini-Batch Training Algorithm}
    \label{fig:min_batch_train}
\end{figure*}
\begin{figure*}
    \centering
    \includegraphics[width=0.95\textwidth]{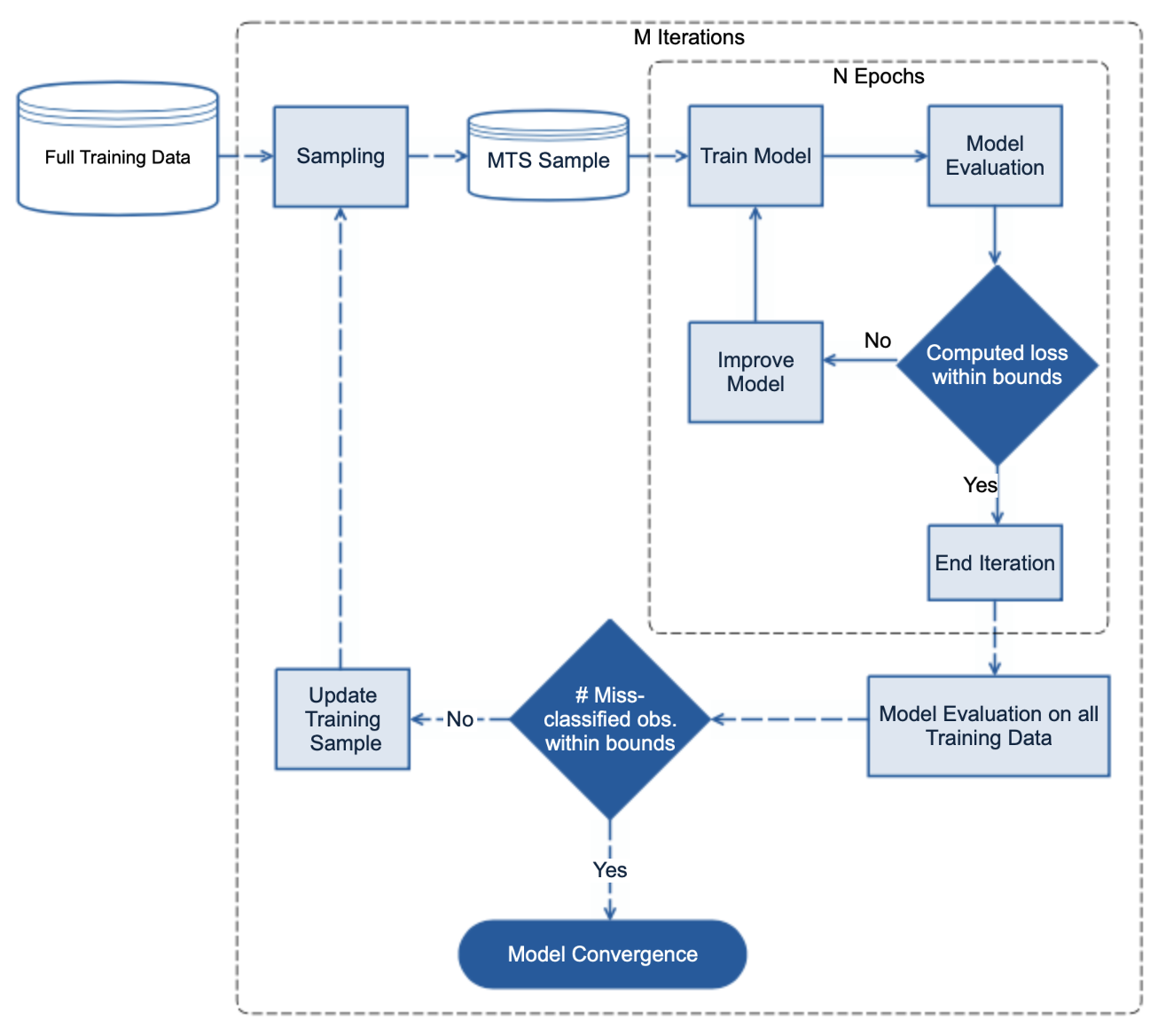}
    \caption{LIIT Training Algorithm}
    \label{fig:LIIT_train}
\end{figure*}

\subsection{LAD Improved Iterative Training of The Neural Network}
Traditionally, in batch training, the full training data is divided into smaller samples or batches. The ANN learns from each batch sequentially till all the observations from the full training data are exhausted, as demonstrated in Figure \ref{fig:min_batch_train}. In the LIIT training, we iteratively design and update the modified training samples, MTS, from the full training data. At each iteration, we train the ANN using batch training on MTS till convergence. This partially trained model is then tested on the full training data to identify potential learning flaws. Since the current work is limited to classification models, the learning flaws include the misclassified data.  The misclassified data is then used to derive the updated MTS which is used to retrain the ANN. The process is illustrated in Figure \ref{fig:LIIT_train}. This is inspired by Boosting techniques \cite{schapire2003boosting} where the subset creation depends on the previous model. However, unlike in boosting setting, we retrain the same ANN. \footnote{The LIIT approach is very similar to batch training within the epoch of a neural network. }

To determine and extract the MTS sample, any sampling algorithm can be used. However, to ensure a good representation, we designed four LAD score based sampling algorithms along with the random sampling approach which is used as a baseline. The following are the sampling strategies used in our analysis:  

\begin{enumerate}
    \item \textbf{LAD Anomaly only (Repeated Entry)}: Observations with the highest anomaly scores in each true class are added to the training batch. Multiple copies of the observation can be added over iterations when the model fails to classify them after numerous re-training. See Algorithm \ref{alg:liit_alg_1}. 
    \item \textbf{LAD  Anomaly + Normal (Unique Entry)}: Equal parts of the high and low anomaly score observations are sampled for each true class. The final training batch contains a unique set of observations with no duplicate entries. See Algorithm \ref{alg:liit_alg_2}.
    \item \textbf{LAD Anomaly only (Unique Entry)}: This is similar to the \textbf{LAD Anomaly only (Repeated Entry)} approach. Observations with the highest anomaly scores in each true class are added to the training batch. However, the final training batch contains a unique set of observations with no duplicate entries. See Algorithm \ref{alg:liit_alg_3}. 
    \item \textbf{LAD Quantile Samples (Repeated Entry)}: The observations are sampled using different quantiles of the anomaly score for each true class. Multiple copies are maintained in the training batch to ensure weighting from under-represented latent classes within each known true class. See Algorithm \ref{alg:liit_alg_4}.  
    \item \textbf{Random}: In this model, we use random sampling from the available data. See Algorithm \ref{alg:liit_alg_5}. 
\end{enumerate}

For this part, we sample $\sim 5-6\%$ of the full training data at each iteration that is later added to the modified training sample. We ensure equal weights for all true classes for the analysis. The LIIT approach is implemented with 6 iterations (1 initial and 5 updates) which brings to $\sim 30\%$ of the full training sample used in the LIIT approach.

\begin{algorithm}
\caption{LAD Anomaly only (Repeated Entry)}
\label{alg:liit_alg_1}
\textbf{Input}:
Dataset $X$ of size $(n,d)$, number of iterations $N_{iter}$, threshold $th$, number of true classes in the data $K$, sample size from each class $c_{size}$, number of iterations $i_{iter}$, ANN classification model $model_{liit}$.\\
\textbf{Initialization}: \\
Split data into $x_{train}, x_{test}, x_{val}, y_{train}, y_{test}, y_{val}$ (train, test and validation)\\
Derive LAD score $ana_{score}$ for all observations in training data i.e. \\ $ana_{score}= LAD(x_{train}, y_{train})$
\begin{algorithmic}[1]
    \State $MTS=[]$ (create empty MTS sample indices list)
    \For{each class $k$}
        \State Generate list of indices of all observations in class $k$, $ind_k$
        \State Subset anomaly scores for each class         \begin{eqnarray*}
            ana_{score_k}=ana_{score}[ind_k]
        \end{eqnarray*}
        \State Identify top $c_{size}$ observations with least anomaly scores and add them samples to the $MTS$ sample i.e. (most non-anomalous observations)
        \For{each iteration $i\leq i_{iter}$}
            \State Fit the ANN on $MTS$ using batch training,
            \begin{eqnarray*}
                model_{liit}.fit(x_{train}[MTS], y_{train}[MTS])
            \end{eqnarray*}
            \State Predict model classification on $x_{train}$, $z_{pred}=model_{liit}.predict(x_{train})$
            \State Identify all miss-classified observations' indices in training data 
            \begin{eqnarray*}
                err_{inds}= np.where(z_{pred}!=y_{train})
            \end{eqnarray*}
            \For{each class $k$}
                \State Identify all miss-classified observations $ind_{err_k}$ 
                \State Subset anomaly scores for miss-classified data in class $k$ 
                \begin{eqnarray*}
                    ana_{err_k}=ana_{score}[ind_{err_k}]
                \end{eqnarray*}
                \State Identify $c_{size}$ observations with highest anomaly scores from $ind_{err_k}$ i.e. (most anomalous observations) and add them to $MTS$ sample.
            \EndFor
        \EndFor
    \EndFor
\end{algorithmic}
\end{algorithm}

\begin{algorithm}
\caption{LAD  Anomaly + Normal (Unique Entry)}
\label{alg:liit_alg_2}
\textbf{Input}:
Dataset $X$ of size $(n,d)$, number of iterations $N_{iter}$, threshold $th$, number of true classes in the data $K$, sample size from each class $c_{size}$, number of iterations $i_{iter}$, ANN classification model $model_{liit}$.\\
\textbf{Initialization}: \\
Split data into $x_{train}, x_{test}, x_{val}, y_{train}, y_{test}, y_{val}$ (train, test and validation)\\
Derive LAD score $ana_{score}$ for all observations in training data i.e. \\ $ana_{score}= LAD(x_{train}, y_{train})$
\begin{algorithmic}[1]
    \State $MTS=[]$ (create empty MTS sample indices list)
    \For{each class $k$}
        \State Generate list of indices of all observations in class $k$, $ind_k$
        \State Subset anomaly scores for each class 
        \begin{eqnarray*}
            ana_{score_k}=ana_{score}[ind_k]
        \end{eqnarray*}
        \State Identify top $c_{size}$ observations with least anomaly scores and add them samples to the $MTS$ sample i.e. (most non-anomalous observations)
        \For{each iteration $i\leq i_{iter}$}
            \State Fit the ANN on $MTS$ using batch training,
            \begin{eqnarray*}
                model_{liit}.fit(x_{train}[MTS], y_{train}[MTS])
            \end{eqnarray*}
            \State Predict model classification on $x_{train}$, $z_{pred}=model_{liit}.predict(x_{train})$
            \State Identify all miss-classified observations' indices in training data 
            \begin{eqnarray*}
                err_{inds}= np.where(z_{pred}!=y_{train})
            \end{eqnarray*}
            \For{each class $k$}
                \State Identify all miss-classified observations $ind_{err_k}$ 
                \State Subset anomaly scores for miss-classified data in class $k$ 
                \begin{eqnarray*}
                    ana_{err_k}=ana_{score}[ind_{err_k}]
                \end{eqnarray*}
                \State Identify $c_{size}/2$ observations each for the lowest and highest anomaly scores from $ind_{err_k}$ i.e. (most anomalous as well as least anomalous observations) and add them to the $MTS$ sample indices.
            \EndFor
            \State Remove repeated indices in the updated modified training sample, 
            \begin{eqnarray*}
                MTS=unique(MTS)
            \end{eqnarray*}
        \EndFor
    \EndFor
\end{algorithmic}
\end{algorithm}

\begin{algorithm}
\caption{LAD Anomaly only (Unique Entry)}
\label{alg:liit_alg_3}
\textbf{Input}:
Dataset $X$ of size $(n,d)$, number of iterations $N_{iter}$, threshold $th$, number of true classes in the data $K$, sample size from each class $c_{size}$, number of iterations $i_{iter}$, ANN classification model $model_{liit}$.\\
\textbf{Initialization}: \\
Split data into $x_{train}, x_{test}, x_{val}, y_{train}, y_{test}, y_{val}$ (train, test and validation)\\
Derive LAD score $ana_{score}$ for all observations in training data i.e. \\ $ana_{score}= LAD(x_{train}, y_{train})$
\begin{algorithmic}[1]
    \State $MTS=[]$ (create empty MTS sample indices list)
    \For{each class $k$}
        \State Generate list of indices of all observations in class $k$, $ind_k$
        \State Subset anomaly scores for each class         \begin{eqnarray*}
            ana_{score_k}=ana_{score}[ind_k]
        \end{eqnarray*}
        \State Identify top $c_{size}$ observations with least anomaly scores and add them samples to the $MTS$ sample i.e. (most non-anomalous observations)
        \For{each iteration $i\leq i_{iter}$}
            \State Fit the ANN on $MTS$ using batch training,
            \begin{eqnarray*}
                model_{liit}.fit(x_{train}[MTS], y_{train}[MTS])
            \end{eqnarray*}
            \State Predict model classification on $x_{train}$, $z_{pred}=model_{liit}.predict(x_{train})$
            \State Identify all miss-classified observations' indices in training data 
            \begin{eqnarray*}
                err_{inds}= np.where(z_{pred}!=y_{train})
            \end{eqnarray*}
            \For{each class $k$}
                \State Identify all miss-classified observations $ind_{err_k}$ 
                \State Subset anomaly scores for miss-classified data in class $k$ 
                \begin{eqnarray*}
                    ana_{err_k}=ana_{score}[ind_{err_k}]
                \end{eqnarray*}
                \State Identify $c_{size}$ observations with highest anomaly scores from $ind_{err_k}$ i.e. (most anomalous observations) and add them to $MTS$ sample.
            \EndFor
            \State Remove repeated indices in the updated modified training sample, 
            \begin{eqnarray*}
                MTS=unique(MTS)
            \end{eqnarray*}
        \EndFor
    \EndFor
\end{algorithmic}
\end{algorithm}

\begin{algorithm}
\caption{LAD Quantile Samples (Repeated Entry)}
\label{alg:liit_alg_4}
\textbf{Input}:
Dataset $X$ of size $(n,d)$, number of iterations $N_{iter}$, threshold $th$, number of true classes in the data $K$, sample size from each class $c_{size}$, number of iterations $i_{iter}$, ANN classification model $model_{liit}$.\\
\textbf{Initialization}: \\
Split data into $x_{train}, x_{test}, x_{val}, y_{train}, y_{test}, y_{val}$ (train, test and validation)\\
Derive LAD score $ana_{score}$ for all observations in training data i.e. \\ $ana_{score}= LAD(x_{train}, y_{train})$
\begin{algorithmic}[1]
    \State $MTS=[]$ (create empty MTS sample indices list)
    \For{each class $k$}
        \State Generate list of indices of all observations in class $k$, $ind_k$
        \State Subset anomaly scores for each class 
        \begin{eqnarray*}
            ana_{score_k}=ana_{score}[ind_k]
        \end{eqnarray*}
        \State Identify top $c_{size}$ observations with least anomaly scores and add them samples to the $MTS$ sample i.e. (most non-anomalous observations)
        \For{each iteration $i\leq i_{iter}$}
            \State Fit the ANN on $MTS$ using batch training,
            \begin{eqnarray*}
                model_{liit}.fit(x_{train}[MTS], y_{train}[MTS])
            \end{eqnarray*}
            \State Predict model classification on $x_{train}$, $z_{pred}=model_{liit}.predict(x_{train})$
            \State Identify all miss-classified observations' indices in training data 
            \begin{eqnarray*}
                err_{inds}= np.where(z_{pred}!=y_{train})
            \end{eqnarray*}
            \For{each class $k$}
                \State Identify all miss-classified observations $ind_{err_k}$ 
                \State Subset anomaly scores for miss-classified data in class $k$ 
                \begin{eqnarray*}
                    ana_{err_k}=ana_{score}[ind_{err_k}]
                \end{eqnarray*}
                \State Identify observations corresponding to  $c_{size}$ quantiles in $ana_{err_k}$ scores and add them to the $MTS$ sample indices.
            \EndFor
        \EndFor
    \EndFor
\end{algorithmic}
\end{algorithm}

\begin{algorithm}
\caption{LAD Anomaly only (Repeated Entry)}
\label{alg:liit_alg_5}
\textbf{Input}:
Dataset $X$ of size $(n,d)$, number of iterations $N_{iter}$, threshold $th$, number of true classes in the data $K$, sample size from each class $c_{size}$, number of iterations $i_{iter}$, ANN classification model $model_{liit}$.\\
\textbf{Initialization}: \\
Split data into $x_{train}, x_{test}, x_{val}, y_{train}, y_{test}, y_{val}$ (train, test and validation)\\
Derive LAD score $ana_{score}$ for all observations in training data i.e. \\ $ana_{score}= LAD(x_{train}, y_{train})$
\begin{algorithmic}[1]
    \State $MTS=[]$ (create empty MTS sample indices list)
    \For{each class $k$}
        \State Generate list of indices of all observations in class $k$, $ind_k$
        \State Subset anomaly scores for each class         \begin{eqnarray*}
            ana_{score_k}=ana_{score}[ind_k]
        \end{eqnarray*}
        \State Randomly sample indices of  $c_{size}$ observations and add them samples to the $MTS$ sample
        \For{each iteration $i\leq i_{iter}$}
            \State Fit the ANN on $MTS$ using batch training,
            \begin{eqnarray*}
                model_{liit}.fit(x_{train}[MTS], y_{train}[MTS])
            \end{eqnarray*}
            \State Predict model classification on $x_{train}$, $z_{pred}=model_{liit}.predict(x_{train})$
            \State Identify all miss-classified observations' indices in training data 
            \begin{eqnarray*}
                err_{inds}= np.where(z_{pred}!=y_{train})
            \end{eqnarray*}
            \For{each class $k$}
                \State Identify all miss-classified observations $ind_{err_k}$ 
                \State Subset anomaly scores for miss-classified data in class $k$ 
                \begin{eqnarray*}
                    ana_{err_k}=ana_{score}[ind_{err_k}]
                \end{eqnarray*}
                \State Randomly sample indices of  $c_{size}$ observations and add them samples to the $MTS$ sample
            \EndFor
        \EndFor
    \EndFor
\end{algorithmic}
\end{algorithm}

%% file: LADNN/definitions.tex
\subsection{Definitions and Terminology}
Before describing the detailed methodology, we list out the terminology and corresponding definitions that are used for this study.

\begin{definition}
\textbf{LAD Score} is the Large deviations Anomaly Detection (LAD) generated anomaly score for each observation in the data.  
\end{definition}

\begin{definition}
\textbf{Full-Training Data} is the available complete training dataset for the ANN. It must be noted that only a subset of the Full Training Data might be used to train the ANN in the LIIT approach. Hence we present a different terminology to differentiate it from the training data. 
\end{definition}


\begin{definition}
\textbf{Batch Training} is the traditional ANN training method using mini-batches of training data. 
\end{definition}

\begin{definition}
\textbf{Modified Training Sample (MTS)} is a smaller sample generated from the training data using a specific sampling algorithm.
\end{definition}

\begin{definition}
\textbf{LAD Improved Iterative Training (LIIT)} is the novel improved batch training approach to train the ANN. 
\end{definition}

%% file: LADNN/results.tex
\section{Experiments}\label{experiments_liit}
In this section, we evaluate the classification performance of the simple neural networks on real data when trained using LAD sub-sampled data. We focus on the performance of the neural networks under different training and sampling settings.

The following experiments have been conducted to study the model:
\begin{enumerate}
    \item Computational Expense: The LIIT trained ANN model's ability to train on a smaller set of training samples and converge faster is compared to the fully trained model. 
    \item Classification Performance: The overall performance of the sub-sampled models on multiple benchmark datasets is studied. For this analysis, we consider Area Under the Curve (AUC) as the performance metric to study classification. 
    \item Stability to Perturbations: Perturbations upto 8$\%$ are added to the test data which is used to study the change in performance in all models. 
\end{enumerate}
To maintain fair comparison, the number of epochs is fixed to a maximum count of 180 for the ANN model trained on the full training data a.k.a. the full model and 30 per iteration of all the LIIT trained ANNs (totaling to 180 epochs for complete training). For each trained ANN, we evaluate performance on 5 independent reruns. The average results are presented for all evaluations.

\subsection{Datasets}
We consider a variety of publicly available benchmark data sets from the UCI-ML repository~\cite{Dua:2017}) (See Table ~\ref{tab:datadesc_nn}) for the experimental evaluation. For training, test and validation, the data was randomly split into 80\%, 10\% and 10\% of the data respectively. 

\begin{table}[htb]
  \centering
      \begin{tabular}{|l|c|c|c|}
        \hline
        Name & $N$ & $d$ & $c$\\
        \hline
        Ecoli  &  336  &  7  &  8  \\ 
        Imgseg  &  2310  &  18  &  7  \\ 
        Skin  &  245057  &  4  &  2  \\ 
        Shuttle  &  58000  &  10  &  2  \\ 
        Wisc  &  699  &  9  &  2  \\ 
        Iono  &  351  &  33  &  2  \\ 
        Zoo  &  101  &  16  &  7  \\ 
        Letter  &  20000  &  16  &  26  \\ 
        Comm And Crime  &  1994  &  102  &  2  \\ 
        Vowel  &  990  &  10  &  11  \\ 
        Fault  &  1941  &  28  &  2  \\ 
        Sonar  &  208  &  60  &  2  \\ 
        Balance-Scale  &  625  &  4  &  3  \\ 
        Pageb  &  5473  &  11  &  2  \\ 
        Spambase  &  4601  &  58  &  2  \\ 
        Wave  &  5000  &  22  &  2  \\ 
        Tae  &  151  &  3  &  3  \\ 
        Thy  &  215  &  5  &  3  \\ 
        Opt Digits  &  5620  &  63  &  2  \\ 
        Concrete  &  1030  &  9  &  2  \\ 
      \hline
      \end{tabular}
    \caption{classification Benchmark Datasets: Description of the benchmark data sets used for evaluation of the classification detection capabilities of the proposed model. $N$ - number of instances, $d$ - number of attributes, $c$ - number of true classes in the data set.}
    \label{tab:datadesc_nn}
\end{table}

\begin{figure*}
     \centering
         \subcaptionbox{}{\includegraphics[width=0.9\textwidth]{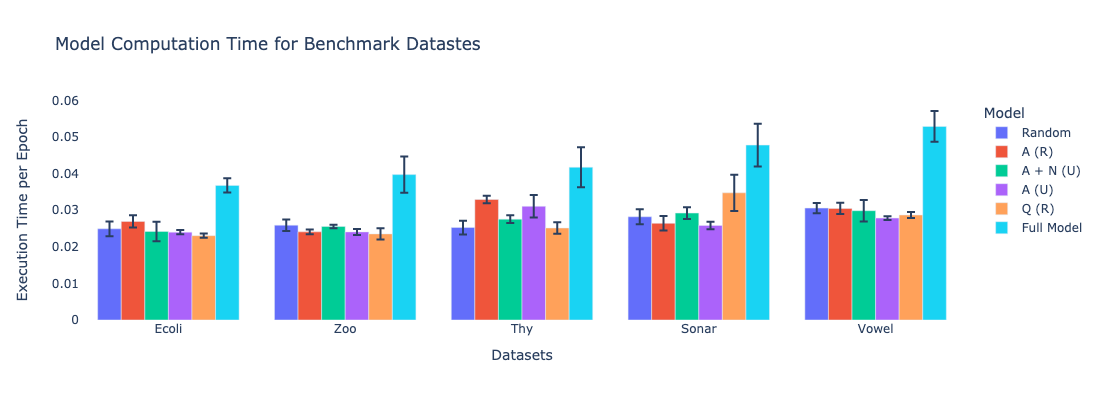}\,
         \label{fig:time_0}}
\end{figure*}
\begin{figure*}\ContinuedFloat 
     \centering
         \subcaptionbox{}{\includegraphics[width=0.9\textwidth]{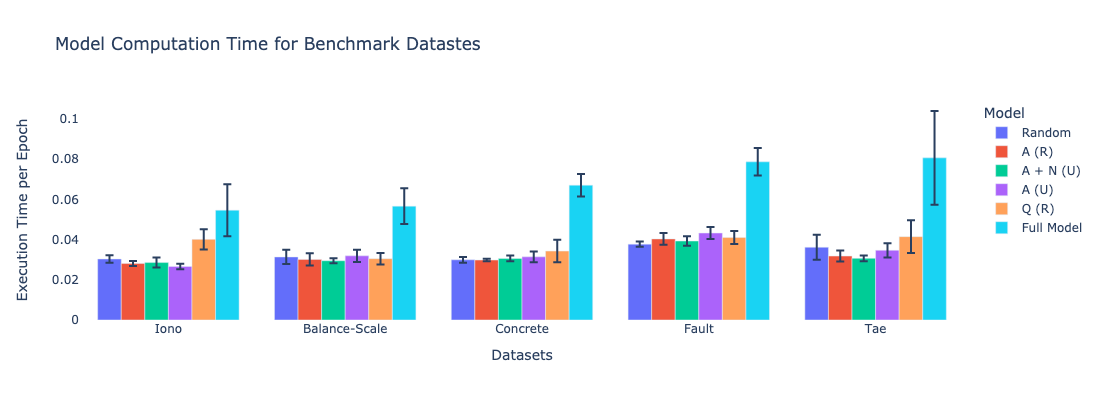}\,
         \label{fig:time_1}}
\end{figure*}
\begin{figure*}\ContinuedFloat 
     \centering
         \subcaptionbox{}{\includegraphics[width=0.9\textwidth]{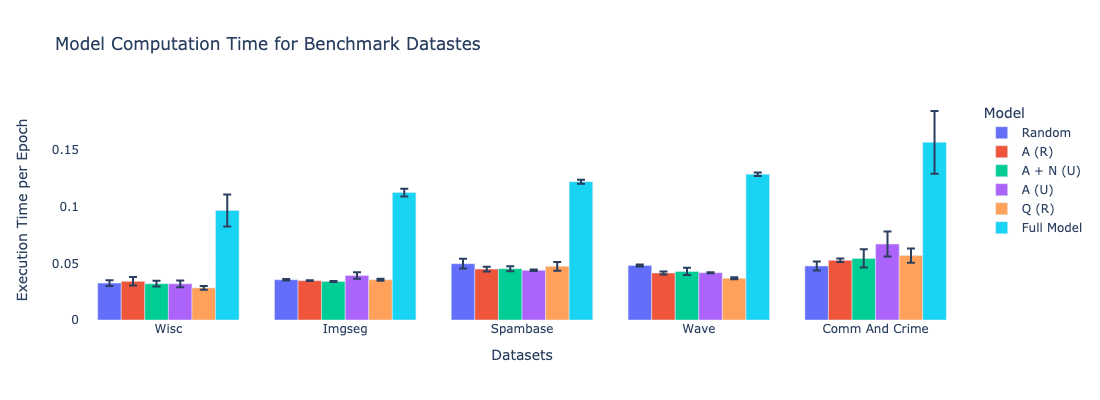}\,
         \label{fig:time_2}}
\end{figure*}
\begin{figure*}\ContinuedFloat 
     \centering
         \subcaptionbox{}{\includegraphics[width=0.9\textwidth]{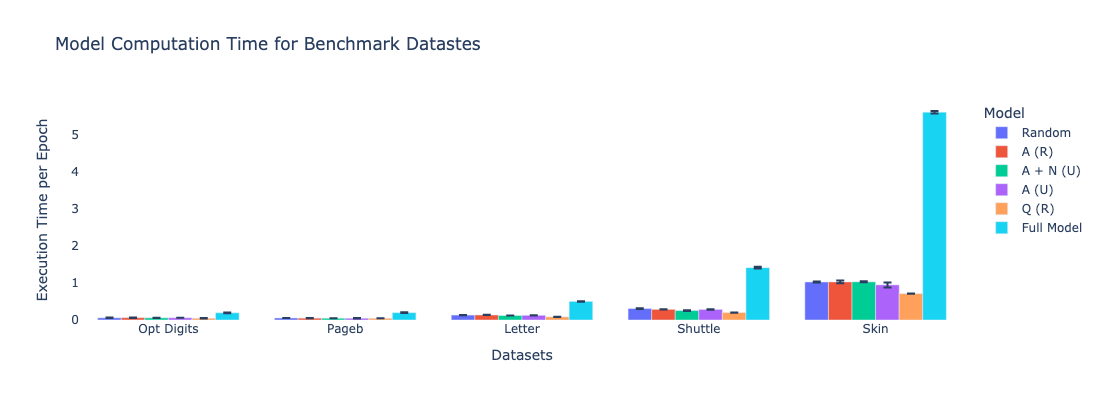} \,
         \label{fig:time_3}}
         \caption{Computation time for different datasets: The figures illustrate the computation time for different LIIT trained ANN models in comparison to the ANN trained on full training data (Full model).}
         \label{fig:liit_com_time_all}
\end{figure*}

\subsubsection{Computational Time}
In this section, we look at the time taken by each ANN to train on the datasets. Since the LIIT trained ANNs use only one-third of the full training data, the training time is evidently lower as compared to training for the full model. This can be clearly seen in the Figures \ref{fig:liit_com_time_all}.

\subsubsection{Model Performance}
Now, since the LIIT trained ANN models have a clear computational advantage over the full model, we look at the overall classification performance on a multitude of benchmark classification datasets. Table \ref{tab:auc_reps} shows the performance of the models on each of these datasets. We use the Area Under the Curve (AUC) as the evaluation metric to study the classification performance of the models. It is discernible that the Quantile Sampling along with LIIT trained ANN model is on par with the fully trained model. 

\begin{sidewaystable}[htb]
  \centering
  \footnotesize
      \begin{tabular}{|l|c|c|c|c|c|c|c|c|}
        \hline
Data & Random & Anomaly (Repeated) & Anomaly + Normal (Unique) & Anomaly (Unique) & Quantile Samples (Repeated) & Full Model \\ 
\hline
Tae & 0.6 ($\pm$ 0.04) & 0.54 ($\pm$ 0.04) & 0.72 ($\pm$ 0.05) & 0.6 ($\pm$ 0.06) & 0.63 ($\pm$ 0.14) & 0.63 ($\pm$ 0.05) \\ 
Spambase & 1.0 ($\pm$ 0.0) & 1.0 ($\pm$ 0.0) & 1.0 ($\pm$ 0.0) & 1.0 ($\pm$ 0.0) & 0.98 ($\pm$ 0.01) & 1.0 ($\pm$ 0.0) \\ 
Comm And Crime & 0.85 ($\pm$ 0.02) & 0.89 ($\pm$ 0.01) & 0.87 ($\pm$ 0.02) & 0.88 ($\pm$ 0.01) & 0.86 ($\pm$ 0.02) & 0.9 ($\pm$ 0.01) \\ 
Wisc & 0.96 ($\pm$ 0.0) & 0.98 ($\pm$ 0.01) & 0.98 ($\pm$ 0.0) & 0.98 ($\pm$ 0.01) & 0.96 ($\pm$ 0.02) & 0.98 ($\pm$ 0.0) \\ 
Letter & 0.99 ($\pm$ 0.0) & 1.0 ($\pm$ 0.0) & 1.0 ($\pm$ 0.0) & 1.0 ($\pm$ 0.0) & 0.97 ($\pm$ 0.0) & 1.0 ($\pm$ 0.0) \\ 
Vowel & 0.94 ($\pm$ 0.01) & 0.91 ($\pm$ 0.01) & 0.87 ($\pm$ 0.02) & 0.91 ($\pm$ 0.01) & 0.75 ($\pm$ 0.05) & 1.0 ($\pm$ 0.0) \\ 
Pageb & 1.0 ($\pm$ 0.0) & 1.0 ($\pm$ 0.0) & 1.0 ($\pm$ 0.0) & 1.0 ($\pm$ 0.0) & 1.0 ($\pm$ 0.0) & 1.0 ($\pm$ 0.0) \\ 
Thy & 1.0 ($\pm$ 0.0) & 0.98 ($\pm$ 0.02) & 1.0 ($\pm$ 0.0) & 1.0 ($\pm$ 0.0) & 0.99 ($\pm$ 0.01) & 1.0 ($\pm$ 0.0) \\ 
Zoo & 0.99 ($\pm$ 0.01) & 1.0 ($\pm$ 0.0) & 0.92 ($\pm$ 0.02) & 1.0 ($\pm$ 0.0) & 0.97 ($\pm$ 0.03) & 1.0 ($\pm$ 0.0) \\ 
Concrete & 0.99 ($\pm$ 0.0) & 0.99 ($\pm$ 0.01) & 0.99 ($\pm$ 0.0) & 0.99 ($\pm$ 0.01) & 0.96 ($\pm$ 0.02) & 0.99 ($\pm$ 0.0) \\ 
Wave & 1.0 ($\pm$ 0.0) & 1.0 ($\pm$ 0.0) & 1.0 ($\pm$ 0.0) & 1.0 ($\pm$ 0.0) & 1.0 ($\pm$ 0.0) & 1.0 ($\pm$ 0.0) \\ 
Fault & 0.98 ($\pm$ 0.01) & 1.0 ($\pm$ 0.0) & 1.0 ($\pm$ 0.0) & 1.0 ($\pm$ 0.0) & 0.92 ($\pm$ 0.01) & 1.0 ($\pm$ 0.0) \\ 
Shuttle & 1.0 ($\pm$ 0.0) & 1.0 ($\pm$ 0.0) & 1.0 ($\pm$ 0.0) & 1.0 ($\pm$ 0.0) & 1.0 ($\pm$ 0.0) & 1.0 ($\pm$ 0.0) \\ 
Opt Digits & 1.0 ($\pm$ 0.0) & 1.0 ($\pm$ 0.0) & 1.0 ($\pm$ 0.0) & 1.0 ($\pm$ 0.0) & 0.99 ($\pm$ 0.0) & 1.0 ($\pm$ 0.0) \\ 
Skin & 1.0 ($\pm$ 0.0) & 1.0 ($\pm$ 0.0) & 1.0 ($\pm$ 0.0) & 1.0 ($\pm$ 0.0) & 1.0 ($\pm$ 0.0) & 1.0 ($\pm$ 0.0) \\ 
Imgseg & 1.0 ($\pm$ 0.0) & 1.0 ($\pm$ 0.0) & 1.0 ($\pm$ 0.0) & 1.0 ($\pm$ 0.0) & 0.99 ($\pm$ 0.01) & 1.0 ($\pm$ 0.0) \\ 
Iono & 0.95 ($\pm$ 0.02) & 0.97 ($\pm$ 0.02) & 0.93 ($\pm$ 0.01) & 0.97 ($\pm$ 0.03) & 0.83 ($\pm$ 0.07) & 0.99 ($\pm$ 0.01) \\ 
Balance-Scale & 0.97 ($\pm$ 0.01) & 1.0 ($\pm$ 0.0) & 0.99 ($\pm$ 0.01) & 0.99 ($\pm$ 0.01) & 0.88 ($\pm$ 0.04) & 1.0 ($\pm$ 0.0) \\ 
Sonar & 0.72 ($\pm$ 0.08) & 0.8 ($\pm$ 0.05) & 0.73 ($\pm$ 0.03) & 0.8 ($\pm$ 0.05) & 0.56 ($\pm$ 0.12) & 0.89 ($\pm$ 0.02) \\ 
Ecoli & 0.95 ($\pm$ 0.01) & 0.92 ($\pm$ 0.03) & 0.85 ($\pm$ 0.03) & 0.95 ($\pm$ 0.02) & 0.92 ($\pm$ 0.02) & 0.95 ($\pm$ 0.0) \\ 
      \hline
      \end{tabular}
    \caption{Model Performance on Test Data: The mean AUC for each model on different datasets is shown here. The standard deviation in AUC on different re-runs is shown here.}
    \label{tab:auc_reps}
\end{sidewaystable}

\subsubsection{Stability to Perturbations}
Since the training samples have a significant influence on the model's learning and performance, we try to look at the stability of the model to various perturbations in the test data. For this, random noise is sampled from a multivariate normal distribution with the $0-8\%$ of the training data mean and variance and is added to all the observations in the test data. Each ANN's performance is evaluated in these settings for all benchmark datasets. 
The final classification performances are seen in Figures \ref{fig:pert_line_all}. It was interesting to note that different datasets had better and relatively more stable performances using different sampling strategies. 



\begin{figure}
         \centering
         \subcaptionbox{}{\includegraphics[width=0.49\textwidth]{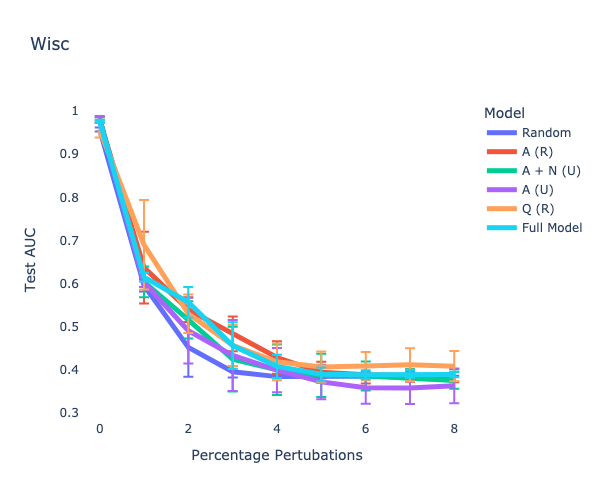}\,
         \label{fig:pert_line_0}}
         \subcaptionbox{}{\includegraphics[width=0.49\textwidth]{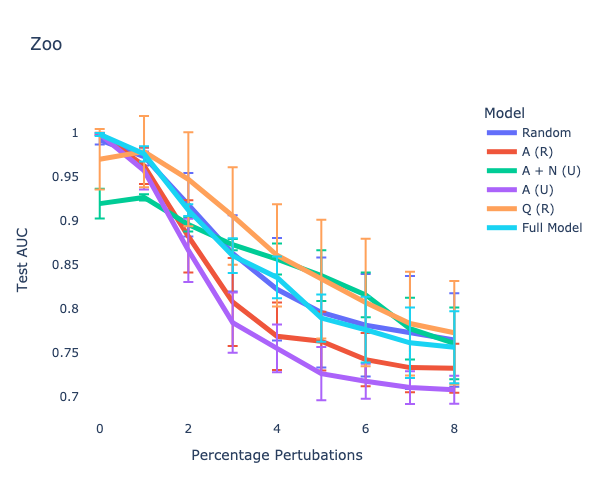}\,
         \label{fig:pert_line_1}}
\end{figure}

\begin{figure}\ContinuedFloat 
         \centering
         \subcaptionbox{}{\includegraphics[width=0.49\textwidth]{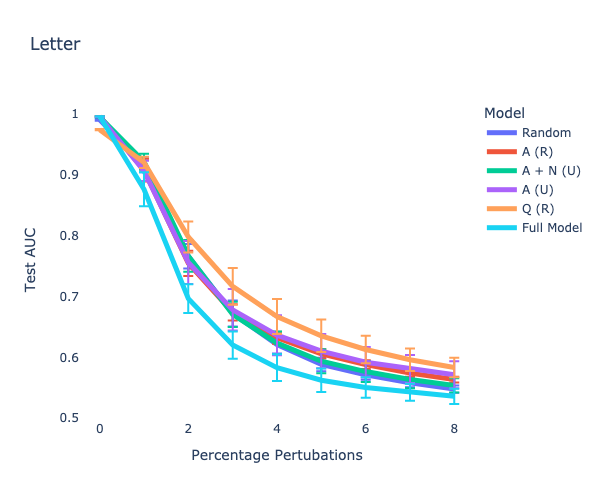}\,
         \label{fig:pert_line_2}}
         \subcaptionbox{}{\includegraphics[width=0.49\textwidth]{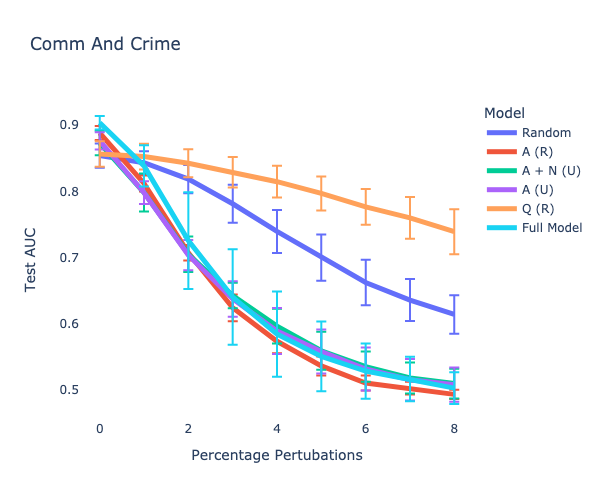}\,
         \label{fig:pert_line_3}}
\end{figure}

\begin{figure}\ContinuedFloat 
         \centering
         \subcaptionbox{}{\includegraphics[width=0.49\textwidth]{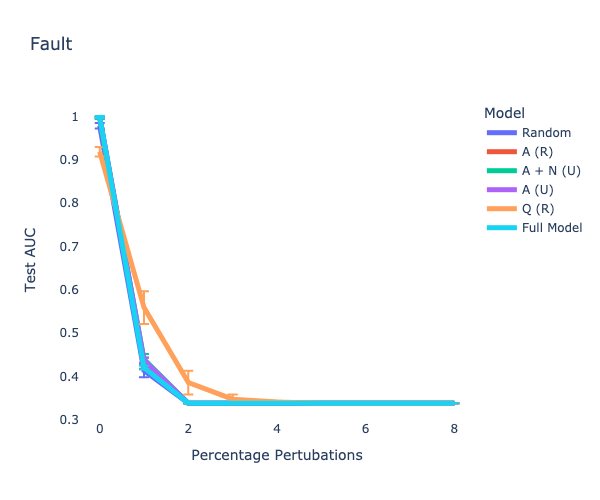}\,
         \label{fig:pert_line_4}}
         \subcaptionbox{}{\includegraphics[width=0.49\textwidth]{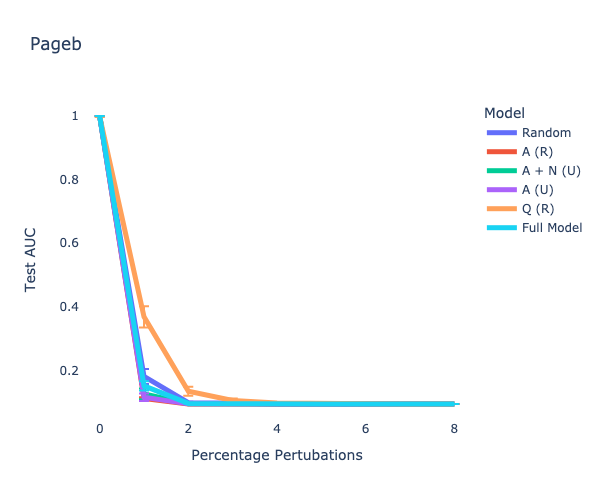}\,
         \label{fig:pert_line_5}}
\end{figure}

\begin{figure}\ContinuedFloat 
         \centering
         \subcaptionbox{}{\includegraphics[width=0.49\textwidth]{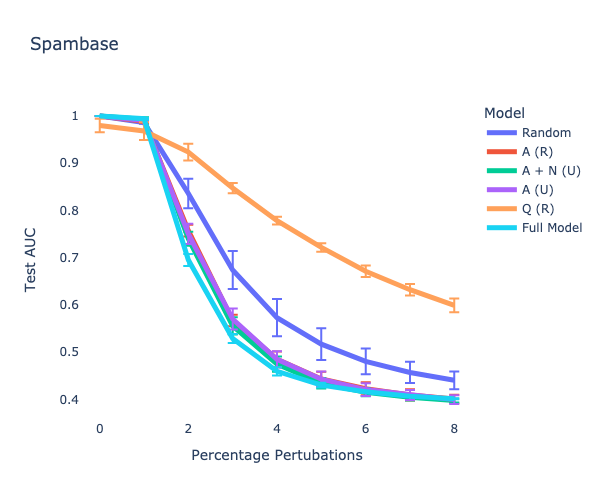}\,
         \label{fig:pert_line_6}}
         \subcaptionbox{}{\includegraphics[width=0.49\textwidth]{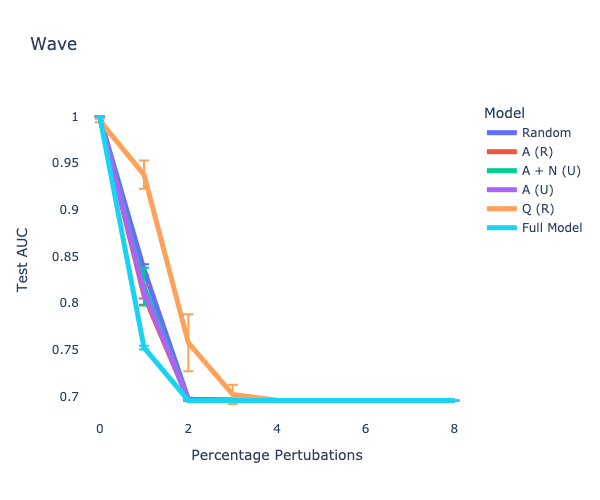}\,
         \label{fig:pert_line_7}}
\end{figure}

\begin{figure}\ContinuedFloat 
         \centering
         \subcaptionbox{}{\includegraphics[width=0.46\textwidth]{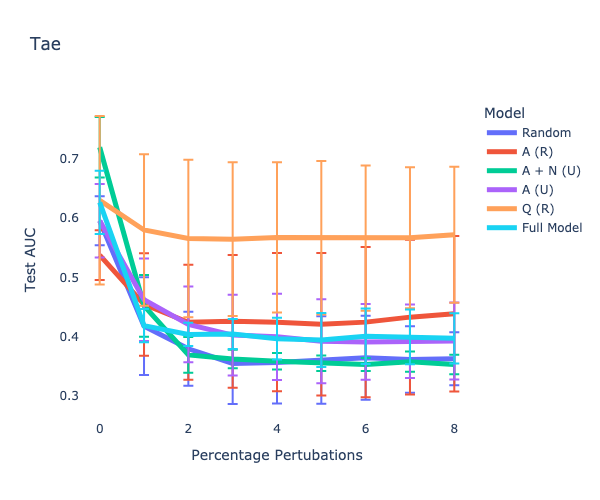}\,
         \label{fig:pert_line_8}}
         \subcaptionbox{}{\includegraphics[width=0.46\textwidth]{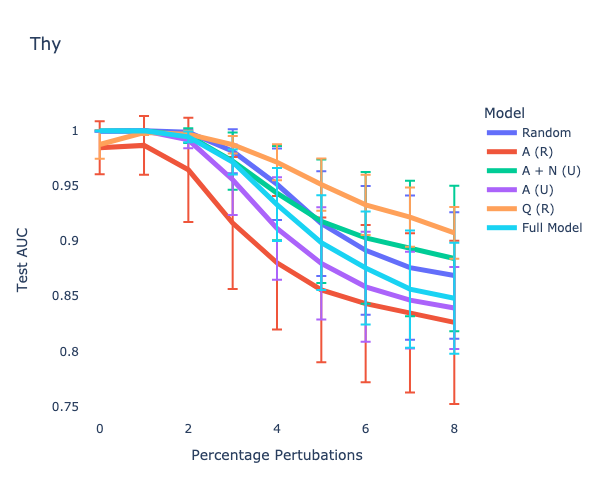}\,
         \label{fig:pert_line_9}}
         \subcaptionbox{}{\includegraphics[width=0.46\textwidth]{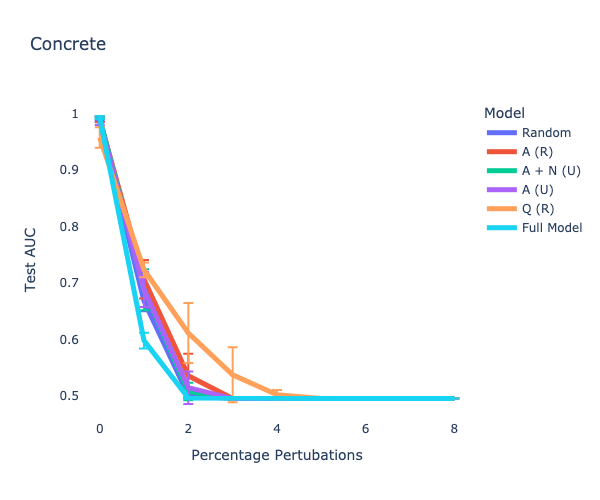}\,
         \label{fig:pert_line_10}}
\end{figure}

\begin{figure}\ContinuedFloat 
         \centering
         \subcaptionbox{}{\includegraphics[width=0.49\textwidth]{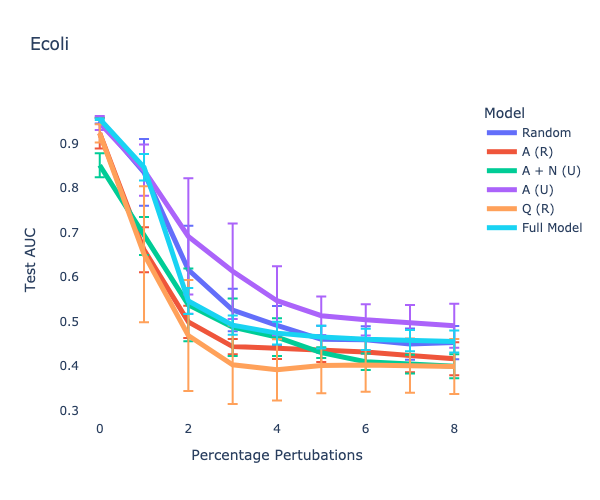}\,
         \label{fig:pert_line_11}}
         \subcaptionbox{}{\includegraphics[width=0.49\textwidth]{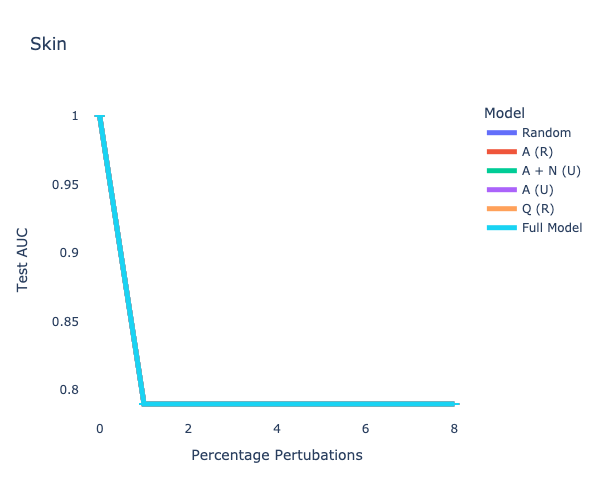}\,
         \label{fig:pert_line_12}}
\end{figure}

\begin{figure}\ContinuedFloat 
         \centering
         \subcaptionbox{}{\includegraphics[width=0.49\textwidth]{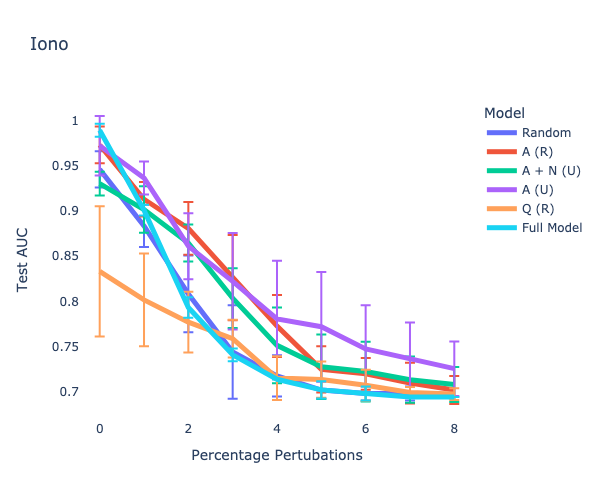}\,
         \label{fig:pert_line_13}}
         \subcaptionbox{}{\includegraphics[width=0.49\textwidth]{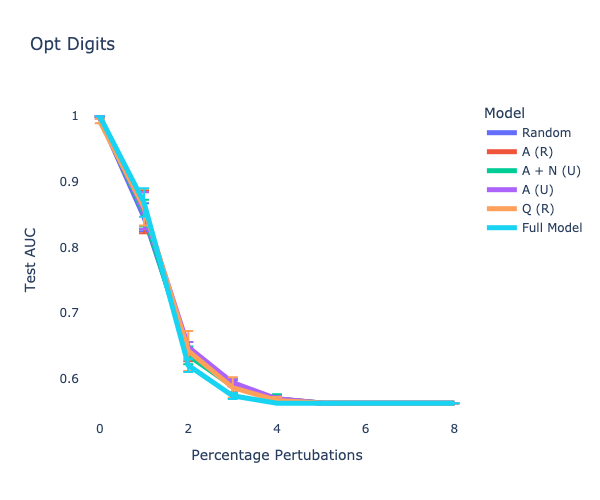}\,
         \label{fig:pert_line_14}}
\end{figure}

\begin{figure}\ContinuedFloat 
         \centering
         \subcaptionbox{}{\includegraphics[width=0.49\textwidth]{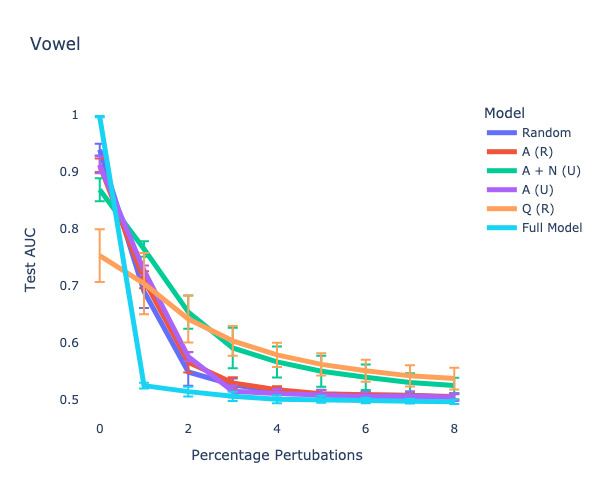}\,
         \label{fig:pert_line_15}}
         \subcaptionbox{}{\includegraphics[width=0.49\textwidth]{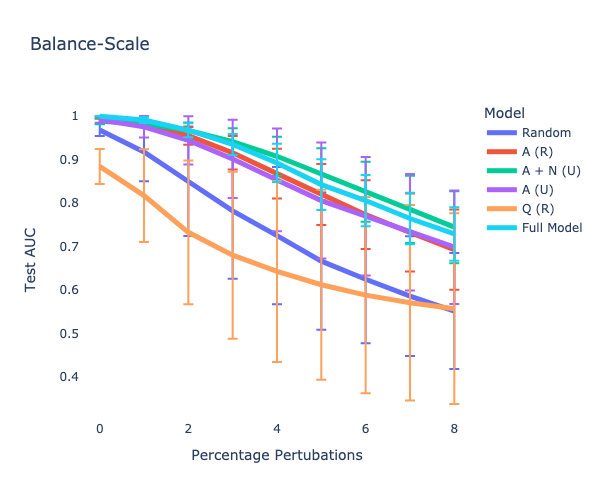}\,
         \label{fig:pert_line_16}}
\end{figure}
\begin{figure}\ContinuedFloat 
         \centering
         \subcaptionbox{}{\includegraphics[width=0.46\textwidth]{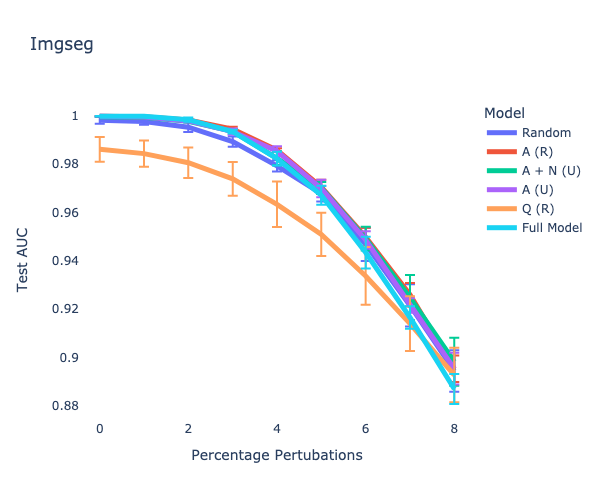}\,
         \label{fig:pert_line_17}}
         \subcaptionbox{}{\includegraphics[width=0.46\textwidth]{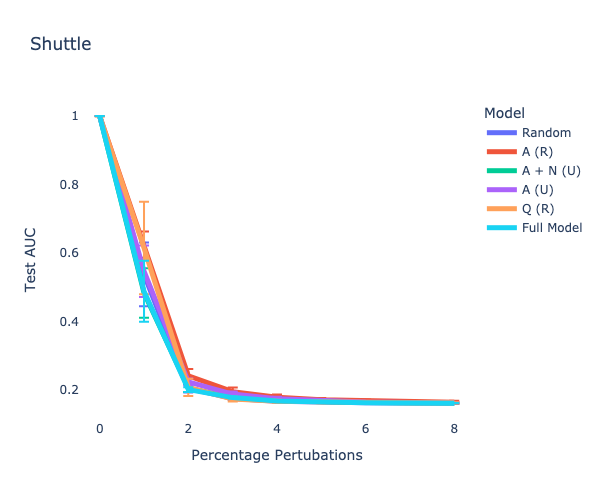}\,
         \label{fig:pert_line_18}}
         \subcaptionbox{}{\includegraphics[width=0.46\textwidth]{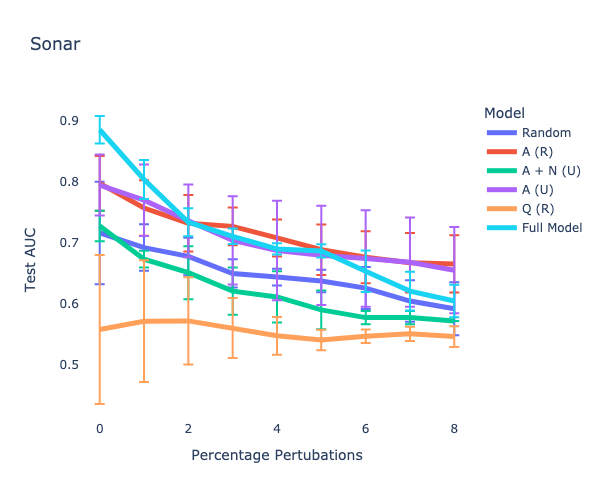}\,
         \label{fig:pert_line_19}}
         \caption{Change in AUC over $\%$ perturbations added to the data: The figure illustrates the change in AUC with increase in $\%$ perturbations to different datasets. The error-bars indicate the standard deviations in AUC values for 5 repetitions}
         \label{fig:pert_line_all}
\end{figure}

Now, to see the individual changes in performance to perturbations, we look at the raw change in AUC values due to the addition of perturbations for all models. Figures \ref{fig:per_all} show the change in performance for different datasets. In particular, Figures \ref{fig:per_0} and \ref{fig:per_1} show a group of datasets that show better performance using Quantile (Repeated), while Figures \ref{fig:per_2} - \ref{fig:per_4} show performance on datasets where Anomaly (Unique), Anomaly + Normal (Repeated) and Anomaly (Repeated) sampling approaches have respectively outperformed.


It can be seen that the Quantile Sample Trained Model has a higher mean AUC as well as lower deviation in AUC than the fully trained model in most datasets. 

\begin{figure}[h!]
         \centering
         \subcaptionbox{Datasets where Quantile Sampling LIIT trained model shows the best performance \label{fig:per_0}}{\includegraphics[width=0.49\textwidth]{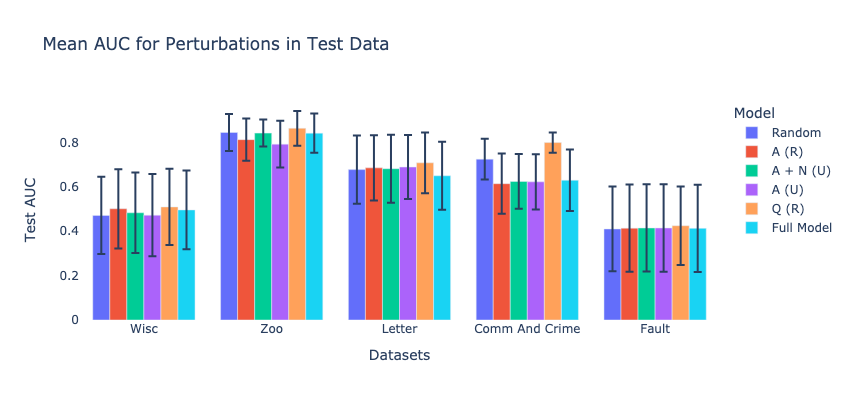} \,
         }
\end{figure}
\begin{figure}[h!]\ContinuedFloat 
         \centering
         \subcaptionbox{{Datasets where Quantile Sampling LIIT trained model shows the best performance}
         \label{fig:per_1}}{\includegraphics[width=0.49\textwidth]{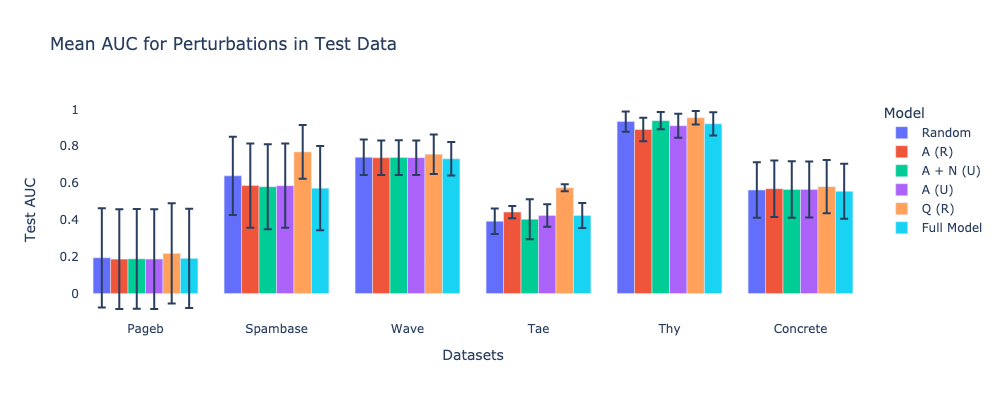} \,
         }
\end{figure}
\begin{figure}[h!]\ContinuedFloat 
         \centering
         \subcaptionbox{Datasets where Anomaly (Unique) LIIT trained model show best performance\label{fig:per_2}}{\includegraphics[width=0.49\textwidth]{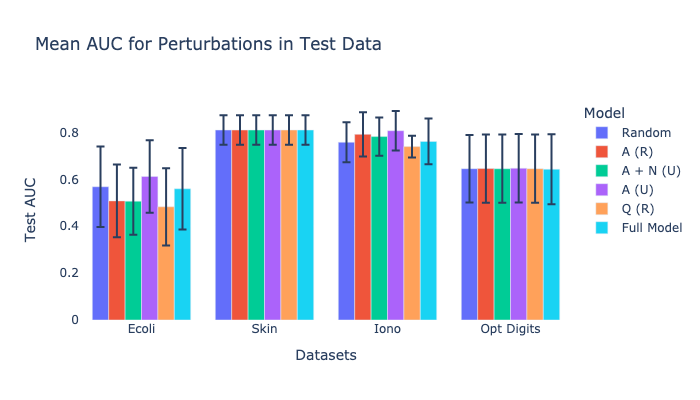}\,
         }
\end{figure}
\begin{figure}[h!]\ContinuedFloat 
         \centering
         \subcaptionbox{Datasets where Anomaly + Normal (Repeated) LIIT trained model show best performance\label{fig:per_3}}{\includegraphics[width=0.49\textwidth]{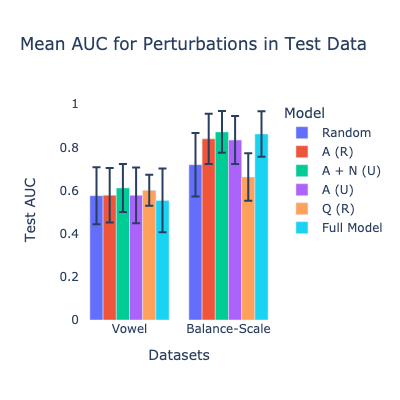}\,
         }
\end{figure}
\begin{figure}[h!]\ContinuedFloat 
         \centering
         \subcaptionbox{Datasets where Anomaly (Repeated) LIIT trained model shows the best performance\label{fig:per_4}}{\includegraphics[width=0.49\textwidth]{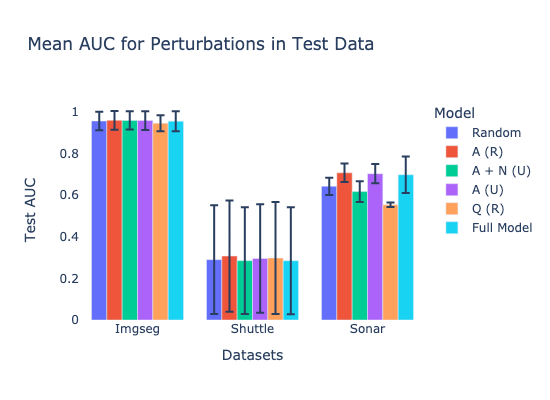}\,
         }
         \caption{Total change in AUC over $\%$ perturbations added to the data: The figure illustrates the mean change in AUC with the increase in $\%$ perturbations to different datasets. The error-bars indicate the standard deviations in AUC values across perturbations.}
         \label{fig:per_all}
\end{figure}

Here, we can see that different LIIT models outperform for different datasets. We hypothesize that the data distribution and heterogeneity play important role in the overall performance and stability. We intend to continue the study of the proposed hypothesis as future research.

\section{Conclusion}
We present a new training strategy for enhancing the learning speed of a neural network whilst maintaining the performance of the model. We present the LAD Improved Iterative Training (LIIT) which is an improved iterative training version of the traditional batch training approach. The LIIT approach uses a modified training sample (MTS) generated and updated using a LAD score based sampling approach that ensures enough representation of extreme and rare behaviours. In particular, the LAD score based Quantile Sampling approach allows ample heterogeneity within the sample data. We study the classification performance of the LIIT trained ANN in comparison with ANN trained on full training data on real benchmark datasets. Though the current research is limited to simple classification neural networks, the work has immense research potential. The LIIT training approach combined with specific LAD sampling methodology might draw out the best performance in a dataset based on the data characteristics. Future studies might help understand the impact of data heterogeneity and sampling method on the performance of ANN. 